\documentclass[twoside]{article}

\usepackage{epsfig}
\usepackage{graphicx}
\usepackage{amsmath}
\usepackage{amssymb}
\usepackage{amsmath} 
\usepackage{mathtools}
\usepackage{graphicx}
\usepackage{multirow}
\usepackage{subfig}
\usepackage{comment}
\newcommand\numberthis{\addtocounter{equation}{1}\tag{\theequation}}
\usepackage{mathtools}
\usepackage[title]{appendix}

\usepackage[accepted]{aistats2022}
\begin{document}

%

%

\twocolumn[

\aistatstitle{Deep Dive into Semi-Supervised ELBO for Improving Classification Performance}

\aistatsauthor{ Fahim Faisal Niloy \And M Ashraful Amin \And AKM Mahbubur Rahman \And Amin Ahsan Ali}

\aistatsaddress{ Agency Lab, Independent University Bangladesh } ]

\begin{abstract}
  Decomposition of the evidence lower bound (ELBO) objective of VAE used for density estimation revealed the deficiency of VAE for representation learning and suggested ways to improve the model. In this paper, by decomposing the ELBO specifically for semi-supervised VAE, we address two problems that were not explored previously. Specifically, we show that mutual information between input and class labels decreases during maximization of semi-supervised ELBO objective. We also point out that semi-supervised VAE deviates from cluster assumption during optimization. We propose methods to address both the issues. Experiments on diverse datasets verify that our method can be used to improve the classification performance of existing VAE based semi-supervised models. Experiments also show that, this can be achieved without sacrificing the generative power of the model.
\end{abstract}

\section{Introduction}


Success of any machine learning model depends on the quality and diversity of the data available to train the model. However, annotation of large-scale datasets is laborious and time-consuming. Consequently, many algorithms have been proposed that make use of both limited labeled data and vast unlabeled data to learn machine learning models. This learning paradigm is termed as semi-supervised learning or SSL. Semi-supervised classification is a significant part of SSL. In particular, given a training dataset that consists of both labeled and unlabeled instances, semi-supervised classification aims to train a classifier from both the labeled and unlabeled data, such that it generalizes well on unseen instances. SSL algorithms are based on several assumptions about the data distribution \cite{chapelle2009semi}. These include cluster assumption, semi-supervised smoothness assumption, etc., which are essential prerequisite for SSL. A plethora of semi-supervised learning methods exist that utilize these assumptions. Amongst the classical approaches Gaussian mixture or hidden Markov models \cite{zhu2005semi}, non-parametric
density models \cite{kemp2004semi}, semi-supervised support vector machines \cite{bennett1999semi}, graph-based methods \cite{zhu2003semi} etc. are popular. However, one important class of models that are particularly helpful for SSL is the generative models. Generative models are an attractive choice for SSL as they implicitly or explicitly model the underlying data distribution and thus can easily incorporate the unlabeled data to learn data representation.  



Nowadays, deep neural networks
play a dominating role in many research areas. Consequently, many researchers have adopted the classic SSL framework and developed novel SSL methods for deep learning settings. Deep semi-supervised learning methods include generative methods, graph-based methods \cite{liao2018graph}, consistency regularized based methods \cite{tarvainen2017mean, miyato2018virtual} 
etc. The two deep generative models that have received most attention are Generative Adversarial Network (GAN) based models \cite{springenberg2015unsupervised,denton2016semi} and Variational Autoencoder (VAE) based models \cite{kingma2014semi,sonderby2016ladder}. VAEs or latent-variable models are particularly useful for semi-supervised learning because it can easily incorporate unlabeled data as VAEs can learn representation of the data distribution in unsupervised manner. It also has the ability to disentangle representations via the configuration of latent variables \cite{siddharth2017learning}. Moreover, VAEs explicitly model the underlying data distribution that makes these models more attractive choices compared to GANs because the latter model the data distribution implicitly. In this paper we focus on VAE based SSL for classification.

VAE used for density estimation optimizes a lower bound (of data likelihood) objective (ELBO) that uses a variational distribution. Decomposition \cite{hoffman2016elbo, mescheder2017adversarial} of ELBO shows different trade-offs that are implicit in optimizing the VAE objective and the path to refinement of the model. For example, inspired by the decomposition, \cite{tomczak2018vae} introduces more flexible prior to improve inference. \cite{zhao2019infovae, rezaabad2020learning} decompose ELBO to improve representation learning of traditional VAEs. \cite{rosca2018distribution} in a similar way shows that VAEs fail to match marginal distributions in both latent and visible space. \cite{esmaeili2019structured} performs a decomposition similar to \cite{hoffman2016elbo} to develop hierarchically factorized VAE. Decomposition of semi-supervised ELBO will also give us more insights into the inner working of VAE used for semi-supervised learning and also may reveal their potential drawbacks. Addressing the drawbacks will improve VAE based semi-supervised learning. 


Kingma et al. \cite{kingma2014semi} have first addressed the problem of semi-supervised classification using variational inference. Subsequent works \cite{maaloe2016auxiliary, sonderby2016ladder} improve upon Kingma's work by introducing auxiliary variables, multiple layer of latent variables etc. Within traditional VAE setting concerning unsupervised density estimation, importance of increasing mutual information between input and latent representation $z$ is studied by several works \cite{zhao2019infovae, rezaabad2020learning}. However, within semi-supervised setting, the mutual information between input and class label $y$ remains to be investigated. In this paper, by decomposing the semi-supervised lower bound objective, we show that mutual information between input and output labels actually decreases during maximization of the unlabeled lower bound (ELBO) objective. This hampers good representation learning. A decreasing mutual information between class representations and inputs is detrimental to the classification objective as well. Also we observe, during maximization, entropy of the classifier increases, which is also harmful for our classification objective as this forces the classifier to deviate from the cluster assumption. To tackle these issues, we regularize the unlabeled lower bound objective so that the mutual information between $y$ and $x$ increases during optimization and also entropy of classifier gets reduced. Various experiments on a wide range of benchmark datasets and prevalent VAE architectures verify that our proposed method can improve the semi-supervised classification accuracy of these VAE models. Moreover, experiments on the datasets reveal that reconstruction and sample generation of VAEs do not deteriorate. That is, our proposed method helps to improve classification performance without sacrificing the generative power of VAEs. In summary, our contributions are:

\begin{itemize}
    \item We address the mutual information reduction problem specifically in the case of semi-supervised VAE, which has not been previously explored.
    
    \item We note semi-supervised VAE inherently deviates from cluster assumption during optimization. In order to solve this problem, we regularize entropy of the classifier.
    
    \item We show that our method improves the classification accuracy of prominent semi-supervised VAE models without affecting the generative power.
\end{itemize}

\section{Related Work}

Generative models recognise the
semi-supervised learning problem as a specialised missing data imputation task for the classification problem \cite{kingma2014semi}. In deep learning era generative adversarial networks \cite{goodfellow2014generative} and variational auto-encoders \cite{kingma2013auto} are popular as deep generative models.

\subsection{Semi-supervised VAE}
Kingma et al. \cite{kingma2013auto} introduces variational auto-encoder (VAE) to efficiently approximate inference and learning with directed probabilistic models whose continuous latent variables and/or parameters have intractable posterior distributions. Later, VAE has been incorporated by \cite{kingma2014semi} to solve semi-supervised learning problem. Their generative semi-supervised model describes the data
as being generated by a latent class variable $y$ in addition to a continuous latent variable $z$. Auxiliary Deep Generative Model \cite{maaloe2016auxiliary} extends Semi-Supervised VAE \cite{kingma2014semi} with auxiliary variables. Auxiliary variables can improve the variational approximation by making the variational distribution more expressive and by facilitating the training of deep generative models with multiple stochastic layers. Ladder VAE \cite{sonderby2016ladder} proposes to improve inference by recursively correcting the generative distribution by a data-dependent approximate
likelihood in a process resembling the Ladder Network \cite{rasmus2015semi}. 

\subsection{MI Maximization in VAE}
There are various methods available that maximize mutual information for the purpose of unsupervised representation learning or semi-supervised learning \cite{hjelm2018learning, kong2019mutual, velivckovic2018deep, xie2021muscle}. However, in this paper, we are specifically concerned with VAE. Reduction of mutual information during VAE training is a well-known phenomenon. Quite a few works have recently addressed the issue.

Authors of \cite{hoffman2016elbo, mescheder2017adversarial} show variational autoencoders inherently suffer from mutual information reduction problem, while maximizing the lower bound objective. \cite{alemi2018fixing} derives variational lower and upper bounds on the mutual information between
the input $x$ and the latent variable $z$. Information autoencoding family \cite{zhao2018information} generalizes the optimization objective of VAE to show that the primal optimization problem optimizes the mutual information between latent and visible variables. To address the mutual information reduction problem of VAE, InfoVAE \cite{zhao2019infovae} increases mutual information between $x$ and $z$ during optimization by explicitly adding the mutual information term. InfoMax-VAE \cite{rezaabad2020learning} improves upon the previous work by facilitating the estimation of mutual information. VMI-VAE \cite{serdega2020vmi} introduces a regularizer that ensures mutual information is maximized during VAE training.

All the notable works mentioned above are concerned with mutual information maximization of VAE between $x$ and continuous latent variable $z$ in the case of unsupervised representation learning. However, mutual information problem is yet to be explored in case of semi-supervised VAE, which introduces $y$ as an additional discrete latent variable. 

\subsection{Entropy Minimization}
Reducing entropy to aid in semi-supervised classification has been performed previously by various works that do not involve semi-supervised VAE. Notable works include \cite{grandvalet2005semi}, which introduces a loss term that minimizes the entropy of $p_{\text{model}}(y | x; \theta)$ for unlabeled data x. “Pseudo-Label” \cite{lee2013pseudo} performs entropy minimization implicitly by constructing one-hot labels from high-confidence predictions on unlabeled data and later using these as training labels. In an attempt to unify the dominant approaches for semi-supervised learning, \cite{berthelot2019mixmatch} reduces entropy in the guessed labels for the unlabeled data.

However, for the case of semi-supervised VAE, we point out in section \ref{sectionapproach} that entropy of the classifier increases. That is why, it becomes essential to regularize the entropy while optimizing VAE objective. This specific concern in the case of semi-supervised VAE has not been addressed before.

\section{Background}
In this section, we describe the background of the VAE model and show ELBO derivation for SSL.
\subsection{Variational Autoencoder}
We first start with a brief review of VAE \cite{kingma2013auto, rezende2014stochastic}. A latent variable generative model defines a joint distribution $p_{\theta}(x,z)$ between a feature space $z \; \epsilon \; \mathcal{Z}$, and the input space $x \; \epsilon \; \mathcal{X}$. We assume that we have a set of observed data
$X=\{x_{1}, ... ,x_{N}\}$ consisting of $N$ i.i.d. samples of $x$. These samples are assumed to have come from a data distribution $q(x)$. We do not have access to the true distribution of $q(x)$. Only samples are available.

We typically aim at maximizing the marginal log-likelihood $\mathbb{E}_{q(x)}\;[\log p_{\theta}(x)]$ with respect to the parameters. However, when the model is parameterized by deep neural networks, direct optimization of the likelihood becomes intractable.

A classic approach \cite{kingma2013auto}
is to define an amortized inference distribution $q_{\phi}(z|x)$ and jointly optimize a lower bound to the log-likelihood:

\begin{align*}
L_{\theta,\phi}(x) &= \mathbb{E}_{q_\phi(z|x)}\log p_{\theta}(x|z) - KL \;( q_{\phi}(z|x) \; ||  \; p_{\theta}(z)) \\
& \leq \log p_{\theta}(x)
\end{align*}

Naturally, $q_{\phi}(z|x)$ is called the encoder
or inference model, while $p_{\theta}(x|z)$ refers to the decoder or generative model. $\theta$ and $\phi$ are the generative and inference model parameters respectively. Usually we assume a simple prior distribution $p(z)$ over features, such as Gaussian or uniform. We further average this lower bound over the data distribution $q(x)$ to obtain the final optimization objective:

\begin{align*}
L_{ELBO} &= \mathbb{E}_{q(x)} \; [L_{\theta,\phi}(x)] \leq \mathbb{E}_{q(x)}\;[\log p_{\theta}(x)]
\end{align*}

The expectations $\mathbb{E}_{q(x)}$ and $\mathbb{E}_{q_{\phi}(z|x)}$ are empirically approximated via sampling, where samples are drawn based on
$x^{i} \; \epsilon \; q(x)$ and $z^{i} \; \epsilon \; q_{\phi}(z|x)$, and the latter is realized via the reparameterization trick \cite{kingma2013auto}.

\subsection{VAE in Semi-supervised Classification}
\label{subsection-semisupervised}
\begin{figure}[t]
\begin{tabular}{cc}
\subfloat[Generative Model]{\includegraphics[width = 1.4in]{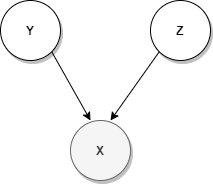}} &
\subfloat[Inference Model]{\includegraphics[width = 1.4in]{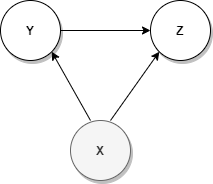}}
\end{tabular}
\caption{Graphical Representation of the M2 model \cite{kingma2014semi}. For a two latent variable VAE for semi-supervised classification, $z$ represents the continous latent variable, $y$ a partially observed latent variable and $x$ represents the observed variable.}
\label{fig:generationM2}
\end{figure}

In semi-supervised setting, let, $X = \{X_{L}, X_{U}\}$ denote the entire data
set, including a small labeled data set $(X_L,Y_L)=\{(x_{1},y_{1}), ... , (x_{n},y_{n})\}.$  and a large scale unlabeled
data set $X_{U} =\{x_{1}, ... , x_{m}\}$, where $m \gg n$. Each observation $x_{i}\; \epsilon \; \mathbb{R}^{D}$ and $y_{i} \; \epsilon \; \{1,...,K\}$. $K$ is the number of categories.

VAEs have been frequently used in semi-supervised classification. Generative semisupervised model, referred to M2 \cite{kingma2014semi}, describes the data
generated by a class variable $y$ and a continuous latent variable $z$. The class labels $y$ are treated as latent variables for unlabeled data. $p(y)$ is assumed as a uniform categorical prior distribution. 
$p_{\theta}(x|z,y)$ is a suitable likelihood function and $q_{\phi}(z|x,y)$ is the inferred posterior distribution. When labels are available the lower bound objective becomes as in \cite{kingma2014semi}:

\begin{align*}
\log p_{\theta}(x,y) \geq &\mathbb{E}_{q_{\phi}(z|x,y)}\; [\log p_{\theta}(x|y,z)+ \log p_{\theta}(y)+ \\ &\log p_{\theta}(z) - \log q_{\phi}(z|x,y)] \\
&= \mathcal{L}(x,y) \numberthis \label{eqn1} 
\end{align*}

When labels are not available $y$ is treated as a discrete latent variable. The unlabeled lower bound objective can be derived in a different manner than given in \cite{kingma2014semi} (detailed proof is given in Appendix):

\begin{align*}
    \log &p_{\theta}(x) = \log \sum_{y}{}\int p_{\theta}(x,y,z)dz \\
    & \geq \mathbb{E}_{q_{\phi}(y,z|x)}\;[\log p_{\theta}(x|z,y)] - KL \;( q_{\phi}(y|x) || p(y) ) \\& - \mathbb{E}_{q_{\phi}(y|x)} \; [KL \;( q_{\phi}(z|x,y) || p(z))] \numberthis \label{eqn2} \\
    &= \mathcal{U}(x)
\end{align*}

Here, $q_{\phi}(y|x)$ is the classifier. 

The unlabeled lower bound objective can also be derived in the same way as \cite{kingma2014semi}: 

\begin{align*}
    \log p_{\theta}(x) & \geq \mathbb{E}_{q_{\phi}(y|x)}\; [\mathbb{E}_{q_{\phi}(z|x,y)} \; [\log p_{\theta}(x|y,z)+ \log p_{\theta}(y) \\& +\log p_{\theta}(z) - \log q_{\phi}(y|x) - \log q_{\phi}(z|x,y)]] \\
    &= \mathbb{E}_{q_{\phi}(y|x)}\; [\mathcal{L}(x,y)] + \mathcal{H}(q_{\phi}(y|x))  \numberthis \label{kingma_lowerbound} \\
    &= \mathcal{U}(x)
\end{align*}

The final objective is:

\begin{align*}
    \mathcal{J} &= \sum_{\mathclap{(x,y) \sim X_{L}}}\mathcal{L}(x,y) + \sum_{\mathclap{x \sim X_{U}}}\mathcal{U}(x) + \alpha\mathbb{E}_{X_{L}}[\log q_{\phi}(y|x)] \numberthis \label{finalobjective}
\end{align*}

We maximize this objective during training.

\section{Approach}
\label{sectionapproach}

One interesting observation arises if we calculate the average of the unlabeled lower bound objective of \eqref{eqn2} over the data distribution. 

\begin{align*}
    \mathbb{E}_{q(x)}[ \log p_{\theta}(x)] &= \mathbb{E}_{q(x)}\;[ \mathbb{E}_{q_{\phi}(z,y|x)}\;[\log p_{\theta}(x|z,y)] \\&- KL \;( q_{\phi}(y|x) || p(y) ) \\&- \mathbb{E}_{q_{\phi}(y|x)} \; [KL \;( q_{\phi}(z|x,y) || p(z))]] \numberthis \label{eqn3}
\end{align*}

We first define two terms:

\begin{align*}
    q_{\phi}(y) &=  \mathbb{E}_{q(x)}\;[q_{\phi}(y|x)]\\
    q_{\phi}(y,x) &= q_{\phi}(y|x) q(x) 
\end{align*}

Here, $q_{\phi}(y)$ is the discrete aggregated posterior \cite{hoffman2016elbo}. The first KL divergence term of \ref{eqn3} can be decomposed as (details in the Appendix):

\begin{align*}
    \mathbb{E}_{q(x)}[KL( q_{\phi}(y|&x) || p(y) )]
    \geq \mathcal{I}_{\phi}(y;x) \numberthis \label{firstmi}
\end{align*}

It shows that the KL divergence is lower bounded by the mutual information between $y$ and $x$. To maximize the unlabeled lower bound objective, KL divergence has to be reduced. As a result, mutual information $\mathcal{I}_{\phi}(y;x)$ also gets decreased during optimization. This is harmful as decreasing mutual information hampers learning of good representations. Also, a decreasing mutual information between class representations and inputs is harmful to classification performance.

Another observation from equation \eqref{kingma_lowerbound} is that, during maximization of lower bound objective entropy of the classifier $\mathcal{H}(q_{\phi}(y|x))$ is increased. Increasing the entropy hurts our classification objective by not conforming to cluster assumption. Cluster assumption states that decision boundary should go through low density regions \cite{chapelle2009semi}. However, increasing the entropy of classifier pushes more data points towards the decision boundary, which in turn deteriorates classifier's performance.

From the above discussion, two possible scopes for improvement are found. We need to somehow regularize the unlabeled lower bound objective such that during optimization mutual information increases and also, the entropy of classifier decreases. For this purpose, we propose to explicitly incorporate both the terms to get the following optimization objective:

\begin{align*}
    \mathcal{M}(x) &= \mathcal{U}(x) + \gamma\;\mathcal{I}_{\phi}(y;x) - \beta\;\mathcal{H}(q_{\phi}(y|x)) \numberthis \label{eqn6}
\end{align*}

Here, $\beta$ and $\gamma$ are hyper-parameters. It should be noted that $\mathcal{M}(x)$ can be thought of as a new optimization objective that regularizes the unlabeled lower bound $\mathcal{U}(x)$ to increase the mutual information and reduce classifier entropy during optimization. Increasing $\mathcal{I}_{\phi}(y;x)$ this way will encourage a better coupling between $y$ and $x$, which helps to learn good representation. Also, a reduced classifier entropy ensures that, the classifier will now push the data points away from decision boundary and help the decision boundary to go through low density regions. Hence, cluster assumption is maintained.

We can calculate $\mathcal{H}(q_{\phi}(y|x))$ in closed form. However, calculation of $\mathcal{I}_{\phi}(y;x)$ is problematic. We now show, mutual information term can be decomposed as:

\begin{align*}
    \mathcal{I}_{\phi}(y;x) &= \int \; \sum_{y}\; q_{\phi}(y,x) \log \frac{q_{\phi}(y,x)}{q_{\phi}(y)q(x)}\;dx \\
    &= - \mathbb{E}_{q(x)}[\mathcal{H}(q_{\phi}(y|x))] + \mathcal{H}(q_{\phi}(y))  \numberthis \label{eqn4}
\end{align*}

Please see Appendix for detailed proof. The expectation in the first term can be estimated with Monte-Carlo estimation using samples from the data distribution. $q_{\phi}(y)$ can also be calculated in a similar way with samples from data distribution and then the entropy of second term is easy to calculate. This way, we can evaluate the mutual information.

Our final objective for maximization is now:

\begin{align*}
    \mathcal{J}_{2} = \sum_{\mathclap{(x,y) \sim X_{L}}}\mathcal{L}(x,y) + \sum_{\mathclap{x \sim X_{U}}}\mathcal{M}(x) + \alpha\mathbb{E}_{X_{L}}[\log q_{\phi}(y|x)] \numberthis \label{eqn7}
\end{align*}

Where the first and last terms are as defined in \eqref{finalobjective}. 

Subsequent works \cite{maaloe2016auxiliary,sonderby2016ladder} on semi-supervised learning extended the setting of M2 model by introducing auxiliary variables, capturing arbitrary dependencies within labels, multiple latent variables, etc. We can similarly add our proposed modifications in the unlabeled lower bound objective of these models. This way, our method can be easily incorporated to existing VAE based semi-supervised models to increase their classification performance.

\section{Experiments}

\subsection{Comparing Methods and Datasets}
We choose three of the most prevalent VAE models for semi-supervised learning as baselines: M2 \cite{kingma2014semi}, Auxiliary Deep Generative Model (ADGM) \cite{maaloe2016auxiliary}, and Ladder Variational Autoencoder (LVAE) \cite{sonderby2016ladder}. We have described M2 in subsection \ref{subsection-semisupervised}. ADGM includes auxiliary variables to improve semi-supervised learning. LVAE enables knowledge sharing between generation and inference model. All three models have distinct mechanisms to improve variational inference and thus help to show the efficacy of our proposed method over a wide variety of variational autoencoders. We change the optimization objective for each of the models to incorporate our proposed modification and refer to the modified models as 'MIER': Mutual Information and Entropy Regularization. We choose four different benchmark datasets: \textit{CIFAR10} \cite{krizhevsky2009learning}, \textit{Fashion-MNIST} \cite{xiao2017/online}, \textit{KMNIST} \cite{clanuwat2018deep}, \textit{SVHN} \cite{netzer2011reading}  for our experiments.

\subsection{Experimental Settings}
The data set for semi-supervised learning is created by splitting the images in training set between a labeled and unlabeled set. The labeled data are chosen randomly and the labeled set is balanced, that is, each class has the same number of labeled images. Rest of the class images in train set are taken as unlabeled set. We keep the same architectural setting for the baseline models as reported in their respective papers \cite{kingma2014semi, maaloe2016auxiliary, sonderby2016ladder}. For M2 model, we use a 50-dimensional latent variable $z$. The encoder and decoder of the inference and generative models are constructed with two fully connected hidden layers, each with 600 units. We use 100-dimensional latent variable $z$ and 100-dimensional auxiliary variable $a$ for ADGM model. Both encoder and decoder have two fully connected hidden layers, each with 500 units. The LVAE has latent variables of dimensions 32, 16 and 8. The encoder and decoder network both have three hidden layers with 128 units. The classifier of all three models consists of one hidden layer with a number of units same as encoder/decoder. We did not use any data augmentation or pre-processing. We converted the color images to grayscale.

For training, we have used the Adam \cite{kingma2014adam} optimization framework with a learning rate of 3e-
4, exponential decay rate for the $1^{st}$ and $2^{nd}$ moment at 0.9
and 0.999, respectively. The learning rate is reduced by half after every 150 epochs. For all the three models, we use a batch size of 200. $\beta$ and $\gamma$ of \eqref{eqn6} is set to 5 and 1 respectively. We have trained all the models for 1000 epochs and saved the best performing model to later report the results on the test set. We have used warm-up \cite{sonderby2016ladder} for all the models, which helps to keep the latent units active during training. The training setting is exactly same for all three models for a fair comparison.

\section{Results}

\begin{table}[t]
\centering
\scriptsize
\caption{Semi-Supervised Classification Accuracy (In $\%$ )} 
\label{tab:classification}
\begin{tabular}{|c|c|c|c|c|}
\hline
Models                    & CIFAR10 & Fashion & SVHN  & KMNIST \\ \hline
M2                        & 25.87   & 78.84        & 28.46 & 65.77  \\ \hline
M2+MIER         & 27.11   & 80.23        & 31.40 & 70.30  \\ \hline
ADGM                      & 26.98   & 79.36        & 26.49 & 67.05  \\ \hline
ADGM+MIER       & 27.82   & 80.19        & 30.98 & 68.97  \\ \hline
LVAE                & 28.14   & 80.48        & 32.04 & 80.48  \\ \hline
LVAE+MIER & \textbf{29.18} & \textbf{80.97} & \textbf{33.25} & \textbf{81.44} \\ \hline
\end{tabular}
\end{table}

\begin{table}[t]
\centering
\caption{Entropy and Negative Log-Likelihood Result} 
\scriptsize
\label{tab:entropy}
\begin{tabular}{|c|c|c|c|c|}
\hline
\multirow{2}{*}{Models} & \multicolumn{2}{c|}{CIFAR10} & \multicolumn{2}{c|}{FashionMNIST} \\ \cline{2-5} 
                        & Entropy       & $\leq logp_{\theta}(x)$          & Entropy         & $\leq logp_{\theta}(x)$             \\ \hline
M2                      & 0.3742        & -636.20      & 0.0439          & -248.18         \\ \hline
M2+MIER                & 0.1551        & -638.31      & 0.0211          & -247.95         \\ \hline
ADGM                    & 0.2920        & -643.66      & 0.0435          & -248.41         \\ \hline
ADGM+MIER              & 0.1435        & -643.42      & 0.0106          & -247.88         \\ \hline
LVAE              & 0.4075        & -652.58      & 0.0770          & -251.35         \\ \hline
LVAE+MIER        & 0.2247        & -652.88      & 0.0452          & -251.51         \\ \hline
\end{tabular}
\end{table}

\subsection{Semi-Supervised Classification}
For our classification experiments, each class has 100 labeled images. We first show the semi-supervised classification results on four benchmark datasets in table \ref{tab:classification}. It shows consistent performance gain over all the datasets and architectures we have tested. The performance gain is more prominent for the M2 model and ADGM with a maximum increase in accuracy of $4.53\%$ and $4.49\%$ respectively. 

\subsection{Entropy of Classifier}
\begin{figure*}[t]
\begin{tabular}{c}
\subfloat[]{\includegraphics[width = 6.5in]{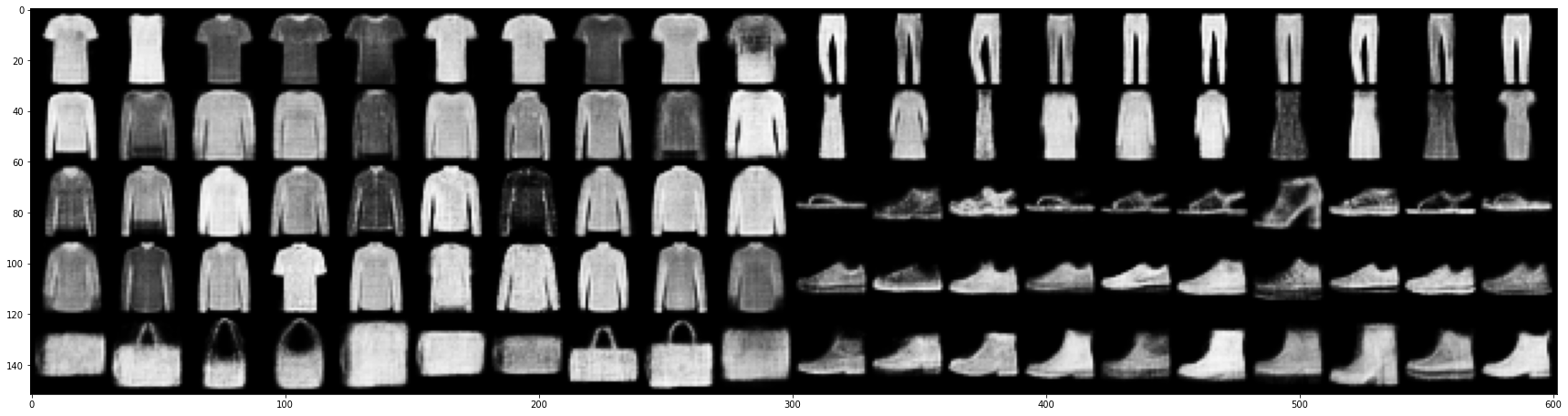}} \\
\subfloat[]{\includegraphics[width = 6.5in]{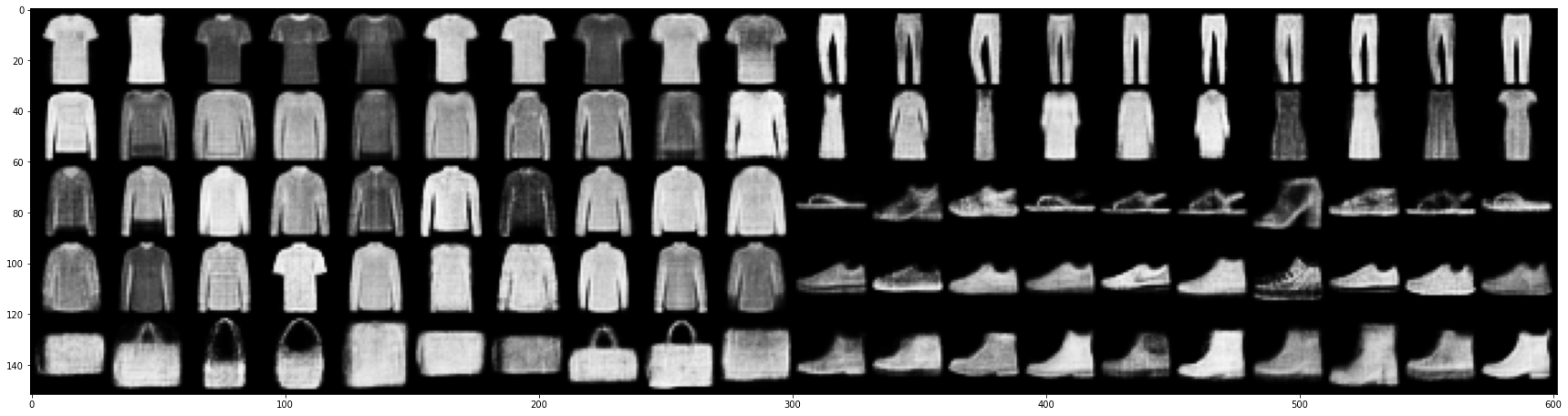}}
\end{tabular}
\caption{(a) Reconstructed samples from the test set of FashionMNIST using M2 model. (b) Sample reconstruction using M2+MIER. Both the images show reconstructed samples are very similar.}
\label{fig:generation}
\end{figure*}

As one of our major objectives is to enforce cluster assumption to improve semi-supervised classification accuracy, we now analyze the changes of classifier entropy with our proposed method over baseline models. We calculate the value of $\mathcal{H}(q_{\phi}(y|x))$ on the test sets of CIFAR10 and FashionMNIST. We choose these two datasets to give a perspective on the value of entropy because classification performance of baseline models is the best on FashionMNIST and least on CIFAR10. The results are given in table \ref{tab:entropy} that shows values of entropy and marginal log-likelihood corresponding to the classification result of table \ref{tab:classification}.

Table \ref{tab:entropy} shows a consistent reduction of entropy with MIER. As entropy value of classifier is high for baseline models on CIFAR10 dataset, the classification accuracy is low. Nevertheless, our method manages to greatly reduce the high entropy over all the three baseline architectures and as a result, classification performance improves. Most importantly, baseline entropy values of classifiers for FashionMNIST are low due to the simpler images of FashionMNIST compared to CIFAR10, still our method manages to reduce the entropy values even further.

\subsection{Generative Log-likelihood Performance}
Table \ref{tab:entropy} also shows the negative log-likelihood (NLL) values for the test sets of CIFAR10 and FashionMNIST. NLL is calculated from the lower bound of equation \eqref{eqn2}. The results suggest that, as the generative network becomes more expressive, the latent codes become less reliant on the inputs \cite{rezaabad2020learning}. Here, it is also observed that, although our method is concerned with increasing mutual information, NLL with our method is very much comparable to NLL of baseline models. Moreover, in some cases (i.e., FashionMNIST with M2 and ADGM, CIFAR10 with ADGM), the values are even better. 

However, getting a comparable NLL while increasing the mutual information may seem to be conflicting. We have shown in \eqref{firstmi} that $\mathcal{I}(y;x)$ arises from $KL( q_{\phi}(y|x) || p(y) )$. As the KL term itself comes as a negative term in the unlabeled lower bound objective of \eqref{eqn2}, in order to maximize marginal log-likelihood, $\mathcal{I}(y;x)$ needs to be minimized. However, our proposed method increases $\mathcal{I}(y;x)$. So, how do we still get a comparable log-likelihood? In order to answer this question, we look at figure \ref{fig:generationM2} of M2 model as an example. Inference is dependent on both latent variable $y$ and $z$. Therefore, a high mutual information between $x$ and $y$ helps to improve inference by learning good representation of $x$, which makes the overall value of equation \eqref{eqn2} high. Consequently, $\log p_{\theta}(x)$ remains similar with baseline. \cite{rezaabad2020learning} finds a similar observation. They increase the mutual information between $x$ and $z$. ELBO decomposition of basic VAE shows this mutual information comes as a negative term in the lower bound objective \cite{hoffman2016elbo}. So, in principle, increasing the mutual information should decrease log-likelihood value in this case. However, they manage to achieve a comparable log-likelihood value over basic VAE. This happens due to the learning of better representation. 

A similar value of $\log p_{\theta}(x)$ indicates that reconstructed images should be quite similar visually. To verify this, we reconstruct images from the test set of FashionMNIST and show the result in Figure \ref{fig:generation}. The top row shows reconstructed images using M2 model trained on 100 labels. The bottom row shows reconstructed images using M2+MIER model. From table \ref{tab:entropy} it is found that value of $\log p_{\theta}(x)$ for both the methods are very similar. As a result, reconstructed images also are very similar visually. The figure experimentally verifies that our method does not sacrifice the generation power of VAEs.


\subsection{Ablation Study for Label Size}

Our proposed method is concerned with the unlabeled semi-supervised ELBO \eqref{eqn2}. So, a natural question arises about how our method performs while increasing the number of labels per class. To answer this question, we perform an ablation study with both M2 and M2+MIER model, to evaluate the model's performance with different number of labeled data made available during training. We opt for 100, 250 and 500 labels per class.

\begin{table}[h]
\centering
\scriptsize
\caption{Ablation Study for Number of Labels (In $\%$ Accuracy)}
\label{table:ablationlabel}
\begin{tabular}{lccc}
             & Number of Labels         & M2    & M2+MIER         \\ \cline{2-4} 
             & \multicolumn{1}{c|}{100} & 25.87 & \textbf{27.11} \\
CIFAR10      & \multicolumn{1}{c|}{250} & 31.08 & \textbf{31.92} \\
             & \multicolumn{1}{c|}{500} & 32.97 & \textbf{33.52} \\ \cline{2-2}
             & \multicolumn{1}{c|}{100} & 78.84 & \textbf{80.23} \\
FashionMNIST & \multicolumn{1}{c|}{250} & 81.97 & \textbf{82.61} \\
             & \multicolumn{1}{c|}{500} & 83.54 & \textbf{84.29} \\ \cline{2-2}
             & \multicolumn{1}{c|}{100} & 28.46 & \textbf{31.40} \\
SVHN         & \multicolumn{1}{c|}{250} & 36.84 & \textbf{37.58} \\
             & \multicolumn{1}{c|}{500} & 41.50 & \textbf{44.06} \\ \cline{2-2}
             & \multicolumn{1}{c|}{100} & 65.77 & \textbf{70.30} \\
KMNIST       & \multicolumn{1}{c|}{250} & 79.11 & \textbf{80.75} \\
             & \multicolumn{1}{c|}{500} & 83.27 & \textbf{83.41} \\ \cline{2-2}
\end{tabular}
\end{table}

\begin{figure*}[t]
\begin{tabular}{cc}
\subfloat[M2]{\includegraphics[width = 3in]{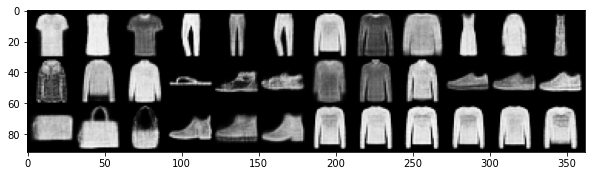}} &
\subfloat[M2 + MIER]{\includegraphics[width = 3in]{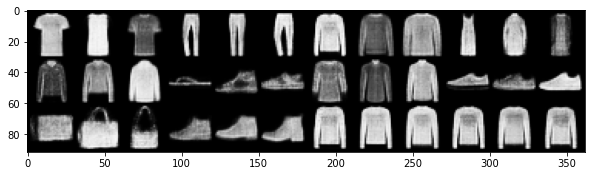}} \\
\subfloat[M2]{\includegraphics[width = 3in]{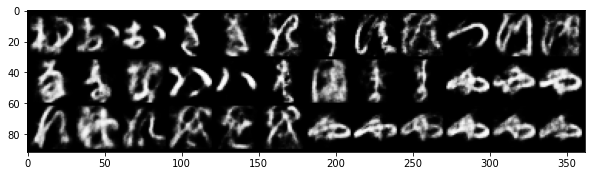}} &
\subfloat[M2 + MIER]{\includegraphics[width = 
3in]{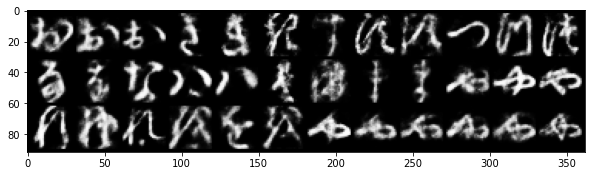}}
\end{tabular}
\caption{Output images generated from noise samples. (a,c): Outputs generated using M2 from CIFAR10 and KMNIST dataset. (b,d): Outputs generated using M2+MIER }
\label{fig:samplegeneration}
\end{figure*}

\begin{figure}[ht!]
\begin{tabular}{cc}
\subfloat[M2 on CIFAR10]{\includegraphics[width = 1.55in]{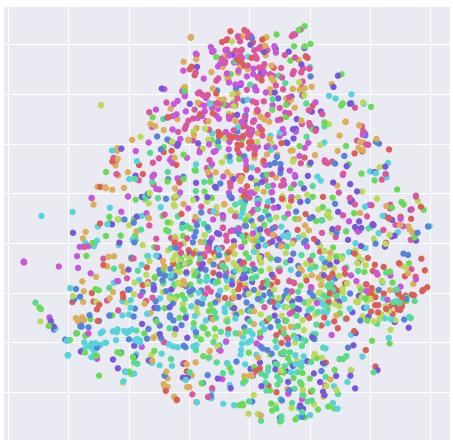}} &
\subfloat[M2+MIER on CIFAR10]{\includegraphics[width = 1.55in]{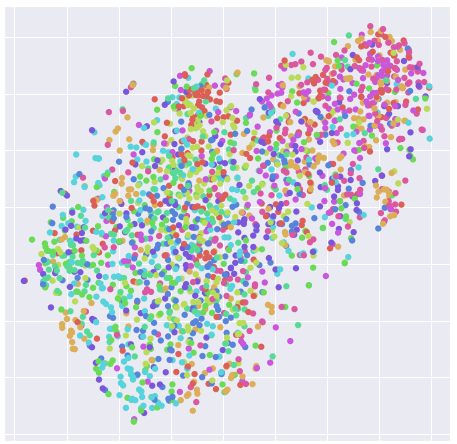}} \\
\subfloat[M2 on FashionMNIST]{\includegraphics[width = 1.55in]{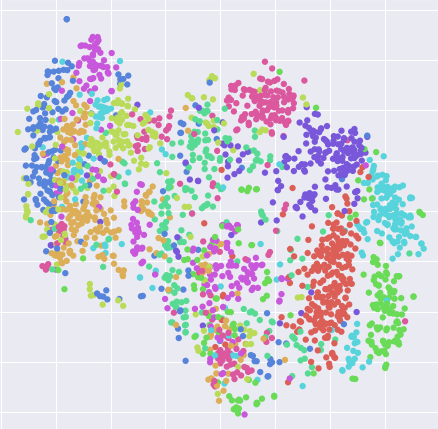}} &
\subfloat[M2+MIER on FashionMNIST]{\includegraphics[width = 1.55in]{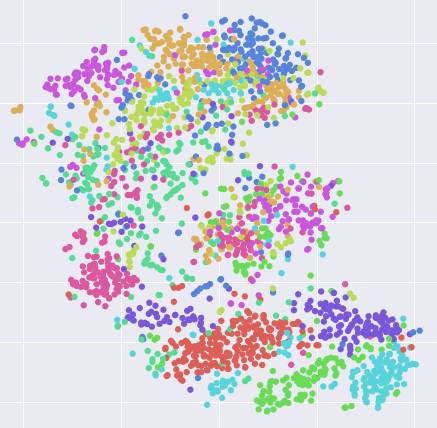}}
\end{tabular}
\caption{(a)-(b): t-SNE diagram of classifier's penultimate layer for M2 model vs M2 + MIER on the test set (2000 points) of CIFAR10 and (c)-(d) FashionMNIST. Both (b) and (d) tend to form more compact clusters than (a) and (c) respectively}
\label{fig:cluster}
\end{figure}

The results are shown in table \ref{table:ablationlabel}. Here also a consistent classification performance gain is observed with M2+MIER model. However, the performance improvement is slightly lower for higher number of labels. The reason is that, with more number of labeled data, the labeled ELBO $\mathcal{L}(x,y)$ of equation \eqref{eqn7} dominates the overall objective. As our method is concerned with unlabeled ELBO, the impact of our method gets slightly reduced with an increasing number of labels. Nevertheless, this can be dealt with by weighting the labeled and unlabeled ELBO of equation \eqref{eqn7}. However, this will increase the number of hyper-parameters. Because in semi-supervised classification, we usually have very small number of labels available per class during training, and even with $500$ labels per class, in our experiment, we have achieved notable classification performance improvement, that is why we have decided not to weight the labelled and unlabeled ELBO of equation \eqref{eqn7}.

\subsection{Cluster Visualization}

In section \ref{sectionapproach}, we argued that our method helps to impose cluster assumption. In summary, cluster assumption states that the decision boundary should go through low density regions \cite{chapelle2009semi}. Decreasing entropy of the classifier helps to achieve this goal. A low value of classifier's entropy over the dataset suggests that the classifier is more confident of its predictions, i.e., a particular class has very high probability and the rest of the classes have low probability. As our proposed method helps in reducing the classifier's entropy, it naturally encourages cluster assumption. However, to verify this visually, we draw the t-SNE diagram of classifier's penultimate layer. Figure \ref{fig:cluster} shows the result.


Top row of the figure corresponds to the t-SNE diagram of (a) M2 model vs (b) M2 model with our method trained with 100 labels per class on CIFAR10 test set. It can be seen that the points are more packed in (b) compared to (a). The bottom row corresponds to the t-SNE diagram of (c) M2 model vs (d) M2 model with our method on FashionMNIST test set trained with the same number of labels. Here it is observed that (d) tends to form more compact clusters than (c). A more clustered feature space translates to higher classification accuracy, which is again verified from table \ref{tab:classification}. Accordingly,  (b) achieves a classification improvement of $1.24\%$ over (a), and (d) achieves a classification improvement of $1.39\%$ over (c).

\subsection{Conditional Generation}
In this section, we visualize the outputs generated from noise samples. We use M2 and M2+MIER to generate three outputs of each class from CIFAR10 and FashionMNIST datasets. We sample a 50 dimensional latent code from $\mathcal{N}(0,$I$)$ and feed it to the decoder. The resulting output images are shown in figure \ref{fig:samplegeneration}. Here also, the sharpness of generated outputs are very similar, which again verifies that our method does not sacrifice VAEs generative power.

\section{Conclusion}
In this paper, we have decomposed the semi-supervised ELBO objective to show that mutual information between input $x$ and class label $y$ decreases during training, which is harmful for the classification objective. To overcome this drawback, we have proposed to regularize the lower bound objective, which involves increasing the mutual information explicitly. We also propose to decrease classifier's entropy which imposes cluster assumption and helps in the overall classification objective. We have performed diverse experiments over four benchmark datasets and three prominent VAE architectures. The experiments verify the efficacy of our proposed method in semi-supervised learning.

\bibliographystyle{plain}
\bibliography{sample_paper}

\onecolumn
\newpage
\aistatstitle{
Supplementary Materials}
\appendixpage
\appendix
\renewcommand{\appendixpagename}{Appendix}
\section{Derivation of Unlabelled Lower Bound Objective \eqref{eqn2}}
\label{FirstAppendix}

We show here a detailed derivation of equation \eqref{eqn2}. Please note from figure \ref{fig:generationM2}:

\begin{align*}
p_{\theta}(x,y,z) &= p_{\theta}(x|z,y)p(y)p(z) \\
q_{\phi}(y,z|x) &= q_{\phi}(z|x,y)q_{\phi}(y|x)
\end{align*}

The log-likelihood of data can be written as:

\begin{align*}
    \log p_{\theta}(x) &= log \; \sum_{y}{}\int p_{\theta}(x,y,z)dz \\
    &= \log \mathbb{E}_{q_{\phi}(y,z|x)}\; [\frac{p_{\theta}(x,y,z)}{q_{\phi}(y,z|x)}]\\
    & \geq \mathbb{E}_{q_{\phi}(y,z|x)}\; log \; [\frac{p_{\theta}(x|z,y)p(y)p(z)}{q_{\phi}(y|x)q_{\phi}(z|x,y)}] \\
    &= \mathbb{E}_{q_{\phi}(y,z|x)}\;[\log p_{\theta}(x|z,y)] - \mathbb{E}_{q_{\phi}(y|x)}\; [\log (\frac{q_{\phi}(y|x)}{p(y)})] - \mathbb{E}_{q_{\phi}(y|x)} \; [\mathbb{E}_{q_{\phi}(z|x,y)} \; \log (\frac{ q_{\phi}(z|x,y)}{p(z)})] \\
     &= \mathbb{E}_{q_{\phi}(y,z|x)}[\log p_{\theta}(x|z,y)] {-} KL( q_{\phi}(y|x) || p(y) )  - \mathbb{E}_{q_{\phi}(y|x)} \; [KL \;( q_{\phi}(z|x,y) || p(z))] \\
    &= \mathcal{U}(x)
\end{align*}

The inequality at third line comes from Jensen's inequality.

\section{Derivation of Mutual Information Term from KL Divergence \eqref{firstmi}}
\label{SecondAppendix}

We now give a detailed derivation of equation \eqref{firstmi}. The data distribution is denoted by $q(x)$.
We also define,
\begin{align*}
    q_{\phi}(y) &=  \mathbb{E}_{q(x)}\;[q_{\phi}(y|x)]\\
    q_{\phi}(y,x) &= q_{\phi}(y|x) q(x) 
\end{align*}

Now,

\begin{align*}
    \mathbb{E}_{q(x)}[KL \;( q_{\phi}(y|x) || p(y) )] &= \int q(x) \sum_{y} \; \log \frac{q_{\phi}(y|x)}{p(y)} \; q_{\phi}(y|x)dx \\
    &= \int q(x)\sum_{y} \; \log \frac{q_{\phi}(y|x) q_{\phi}(y)}{p(y) q_{\phi}(y)}\;q_{\phi}(y|x) dx \\
    &= \int q(x) \sum_{y} \; \log \frac{q_{\phi}(y|x)}{q_{\phi}(y)} q_{\phi}(y|x) dx + \int q(x) \sum_{y} \; \log \frac{q_{\phi}(y)}{p(y)} q_{\phi}(y|x) dx \\ 
    &= \int \; \sum_{y}\; q_{\phi}(y,x) \log \frac{q_{\phi}(y|x)}{q_{\phi}(y)}\;dx + \sum_{y} \; \log \frac{q_{\phi}(y)}{p(y)} q_{\phi}(y) \\
    &= \int \; \sum_{y}\; q_{\phi}(y,x) \log \frac{q_{\phi}(y|x)q(x)}{q_{\phi}(y)q(x)}\;dx + KL \; (q_{\phi}(y) || p(y)) \\
    &= \int \; \sum_{y}\; q_{\phi}(y,x) \log  \frac{q_{\phi}(y,x)}{q_{\phi}(y)q(x)}\;dx + KL \; (q_{\phi}(y) || p(y)) \\
    &= \mathcal{I}_{\phi}(y;x) + KL \; (q_{\phi}(y) || p(y)) \\
    &\geq \mathcal{I}_{\phi}(y;x)
\end{align*}

\section{Decomposition of Mutual Information \eqref{eqn4}}

Mutual Information of equation \eqref{firstmi} can be further decomposed as:

\begin{align*}
    \mathcal{I}_{\phi}(y;x) &= \int \; \sum_{y}\; q_{\phi}(y,x) \log \frac{q_{\phi}(y,x)}{q_{\phi}(y)q(x)}\;dx \\
    &= \int \; \sum_{y}\; q_{\phi}(y|x)q(x) \log \frac{q_{\phi}(y|x)}{q_{\phi}(y)} \; dx \\
    &= \int\; \sum_{y} \; q_{\phi}(y|x)q(x) \log q_{\phi}(y|x) \; dx + \int\; \sum_{y} \; q_{\phi}(y|x)q(x) \log \frac{1}{q_{\phi}(y)} \; dx \\
    &= \int\; q(x) \; \sum_{y} \; q_{\phi}(y|x) \log q_{\phi}(y|x)\; dx - \sum_{y} \; q_{\phi}(y) \log q_{\phi}(y) \\
    &= - \mathbb{E}_{q(x)}[\mathcal{H}(q_{\phi}(y|x)] + \mathcal{H}(q_{\phi}(y)) \\
\end{align*}

\end{document}